\documentclass{article}

\pdfoutput=1

\usepackage{spconf,amsmath,graphicx,color}
\usepackage{siunitx}
\usepackage{booktabs}

\ninept

\title{Generating multilingual voices using speaker space translation based on bilingual speaker data}

\name{Soumi Maiti\sthanks{The first author performed the work while at Apple} and Erik Marchi and Alistair Conkie}
\address{Apple \\
	{\textsuperscript *} The Graduate Center, CUNY, New York, NY}

\begin{document}
\maketitle
\begin{abstract}

We present progress towards bilingual Text-to-Speech which is
able to transform a monolingual voice to speak a second
language while preserving speaker voice quality.  We
demonstrate that a bilingual speaker embedding space contains
a separate distribution for each language and that a simple
transform in speaker space generated by the speaker embedding
can be used to control the degree of accent of a synthetic
voice in a language.  The same transform can be applied even
to monolingual speakers.

In our experiments speaker data from an English-Spanish
(Mexican) bilingual speaker was used, and the goal was to enable
English speakers to speak Spanish and Spanish speakers to
speak English.
We found that the simple transform was sufficient to convert a
voice from one language to the other with a high degree of
naturalness. In one case the transformed voice outperformed a
native language voice in listening tests. Experiments further
indicated that the transform preserved many of the
characteristics of the original voice. The degree of accent
present can be controlled and naturalness is relatively
consistent across a range of accent values.
\end{abstract}


\section{Introduction}
\label{sec:intro}
A European survey found that $56\%$ of people can converse in more than one language \cite{eurobarometer2006europeans}. 
Many people in India, China, and African countries often speak more than one language.
A person is \emph{bilingual} if he or she is able to speak two languages fluently. A challenging problem for text-to-speech (TTS) systems is to make a synthesized  voice bilingual or multi-lingual when recorded speech from only one language is available. A bilingual TTS voice is especially important for pronouncing words from one language embedded within another, i.e., \emph{code switching}.

TTS voices are traditionally synthesized with monolingual
speech. Changing the language of the text may result in a
change of voice, which can be disconcerting or confusing to the listener.
In contrast, it is
possible to use hand-crafted mapping of phonemes from one
language to another, which may not change the voice
characteristics but results in heavily accented speech. The
challenge for bilingual TTS is to maintain the same
speaker voice characteristics while maintaining fluency or
naturalness in both languages.

There are two areas
of work related to this paper. The first utilizes data from a
bilingual speaker to create a bilingual TTS system. In
\cite{ming2017light, fan2016speaker}, the authors model
English-Chinese TTS with data from one and three bilingual
speakers, in \cite{fan2016speaker} one particular case of
cross-lingual transfer is shown. Both models achieve
relatively low speech naturalness in terms of Mean Opinion
Score (MOS), with MOS $\leq 3$.  In the second area of work,
multilingual TTS is trained on a large number of speakers. In
\cite{rallabandi2017building} the authors observe partial
sharing of similar sounding phonemes is preferred compared to
separate phonemes in multiple Indian-English code switched
sentences. In \cite{nachmani2019unsupervised}, an
English-Spanish-German multilingual TTS model is trained with
409 speakers. The authors introduce a speaker-preserving loss
to improve speaker similarity while converting a voice to
another language. However, since they use the non-neural WORLD
\cite{morise2016world} vocoder, naturalness of voices is low
(MOS $\sim 3$). In another recent paper
\cite{zhang2019learning}, the authors built a multilingual TTS
with 550 hours of data from 92 speakers. They show that, when
trained with a large number of speakers, cross-lingual transfer
with high naturalness is possible with a loss of speaker
similarity.

In contrast to the work listed above, our TTS model is trained on a
smaller dataset (56 hours and 7 speakers) and can maintain
speaker similarity close to a bilingual speaker. We show that we can
use information from the speaker embedding space to transform a monolingual
voice to another language thus creating a true bilingual TTS.
The key contributions of this paper are as follows:
We propose a bilingual speaker TTS model,
i.e. one TTS voice that speaks both languages with
high naturalness. We also show that bilingual speaker embedding
space contains separate distributions for two
languages with visualization with Principal Component Analysis (PCA) and by fitting
two Gaussian distributions with Linear Discriminant Analysis (LDA) with 99\%
accuracy. We demonstrate that monolingual TTS voices can made to speak
a second language fluently with a fixed shift
($\Delta$) in the speaker embedding vector. This
also gives us a means of controlling the degree of accent present in
cross-lingual transfer.

\section{Technical Overview}\label{sec:tech_ovw}
This work focuses on cross-lingual transfer using English and Spanish as our example languages.
We use Tacotron \cite{wang2017tacotron}
\cite{shen2018natural} to generate a mel-spectrogram from text
and WaveRNN \cite{kalchbrenner2018efficient} to synthesize
speech from the mel-spectrogram. The Tacotron is conditioned on
a speaker embedding d-vector \cite{Heigold2016sv} from the
speaker encoder, trained separately following
\cite{jia2018tarnsfer}.

\subsection{Tacotron-ML}
The baseline Tacotron is a sequence-to-sequence model with
attention that takes phonemes as input and generates
mel-spectrograms as output. Phonemes are converted to 512
dimensional embeddings through an embedding layer and then
processed with a text encoder to generate encoded features of
size 1024. Encoded features are passed to an attention layer
and then an autoregressive decoder to synthesize
mel-spectrograms. With the encoded features we append a
128-dimensional speaker embedding, learned from the speaker
encoder for a multispeaker Tacotron. We add an extra one-hot
Language ID (LID) that converts to a 32-dimensional language
embedding with an embedding layer. The LID for our
case is $0$ for English and $1$ for Spanish.

\begin{figure}[!t]
	\centering
	\includegraphics[width=0.8\linewidth,  trim={0.5cm 0.3cm 0.3cm 0.5cm},clip]{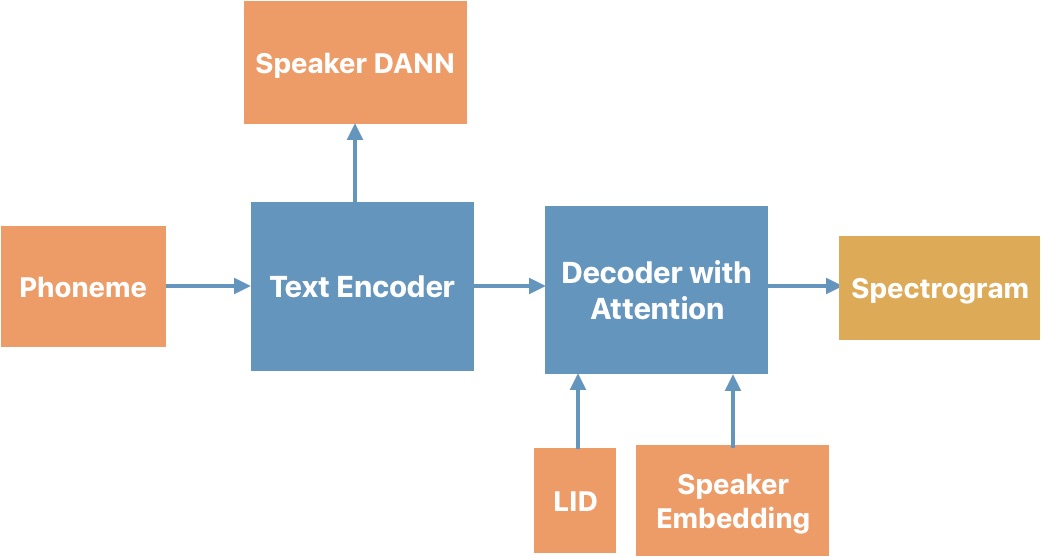}
	\caption{Modified Tacotron-ML}
	\label{fig:model}
\end{figure}

Following \cite{zhang2019learning}, we also add a domain
adversarial neural network (DANN) \cite{ganin2016domain} to
learn speaker-independent encoded features. The speaker DANN
model takes encoded features as input and is trained using a
speaker classification loss. The gradient reversal layer in
the speaker DANN passes the negative gradient scaled by
$\lambda$ to the text encoder during back-propagation. This
helps the text encoder avoid learning speaker information and
instead rely on speaker embeddings for speaker information. In
informal evaluations we found that without a speaker DANN, the
model sometimes generates accented speech, whereas with DANN
the Tacotron produces more consistent results. The modified
Tacotron for multi-lingual (Tacotron-ML) is shown in Figure
\ref{fig:model}. 
During training of the Tacotron-ML, phoneme and LID are computed from the text and the language of the text.
A speaker embedding for each training utterances is obtained from the speaker encoder.
At synthesis time LID and phonemes are extracted from the text to be synthesized, and mean speaker
embedding from training utterances is used for representing the target speaker.
The Tacotron was trained for 1.5M steps with a batch size of 12.

We train a single WaveRNN model with multiple speakers data to synthesize speech from mel-spectrograms, with a GRU~\cite{cho2014learning} size of 512
and training for 800K steps with batch size of 8. A better quality
model might use a single speaker WaveRNN for each speaker, but
the multispeaker WaveRNN is convenient for testing. The single model is also useful for
comparing different speaker results with the same decoder.

\subsection{Speaker Encoder}
The speaker encoder comprises two LSTM~\cite{hochreiter1997long} layers each with 512
units, followed by a 128-unit linear layer and a softmax layer
with 22k units. The input to the LSTM is simply the sequence
of MFCC frames (20 MFCCs per frame, 25ms data window, 100
frames per second). For stochastic optimization, we use Adam
\cite{Kingma14-ADAM} with an initial learning rate of $10^{-3}$ and
a mini-batch size of 128. The speaker encoder is trained
following a curriculum learning procedure which improves both
the robustness against various acoustic conditions and the
generalizability towards less constrained-text scenarios. More
details can be found in \cite{Marchi18-GDT}. In \cite{Hu2019} it was
found useful for in-language voice adaptation.

\begin{table}[bt]
	\centering
	\footnotesize
	\begin{tabular}{l c c c }
		\hline
		Voice ID & Language  & Gen & Locale \\
		\hline
		0 ($ref_{en}$) & en & M & US \\ %
		1              & en & F & US \\ %
		2 ($ref_{es}$) & es & M & MX \\ %
		3              & es & F & MX \\ %
		4              & en & M & AU \\ %
		5              & en & F & AU \\ %
		6              & es & M & ES \\ %
		7              & es & F & ES \\ %
		\hline
	\end{tabular}
	\caption{7 speaker English-Spanish dataset.}
	\label{tab:8sp}
\end{table}

\begin{figure}[!t]
	\centering
	\includegraphics[width=0.35\linewidth, trim={3cm 3cm 3cm 3.5cm},clip]{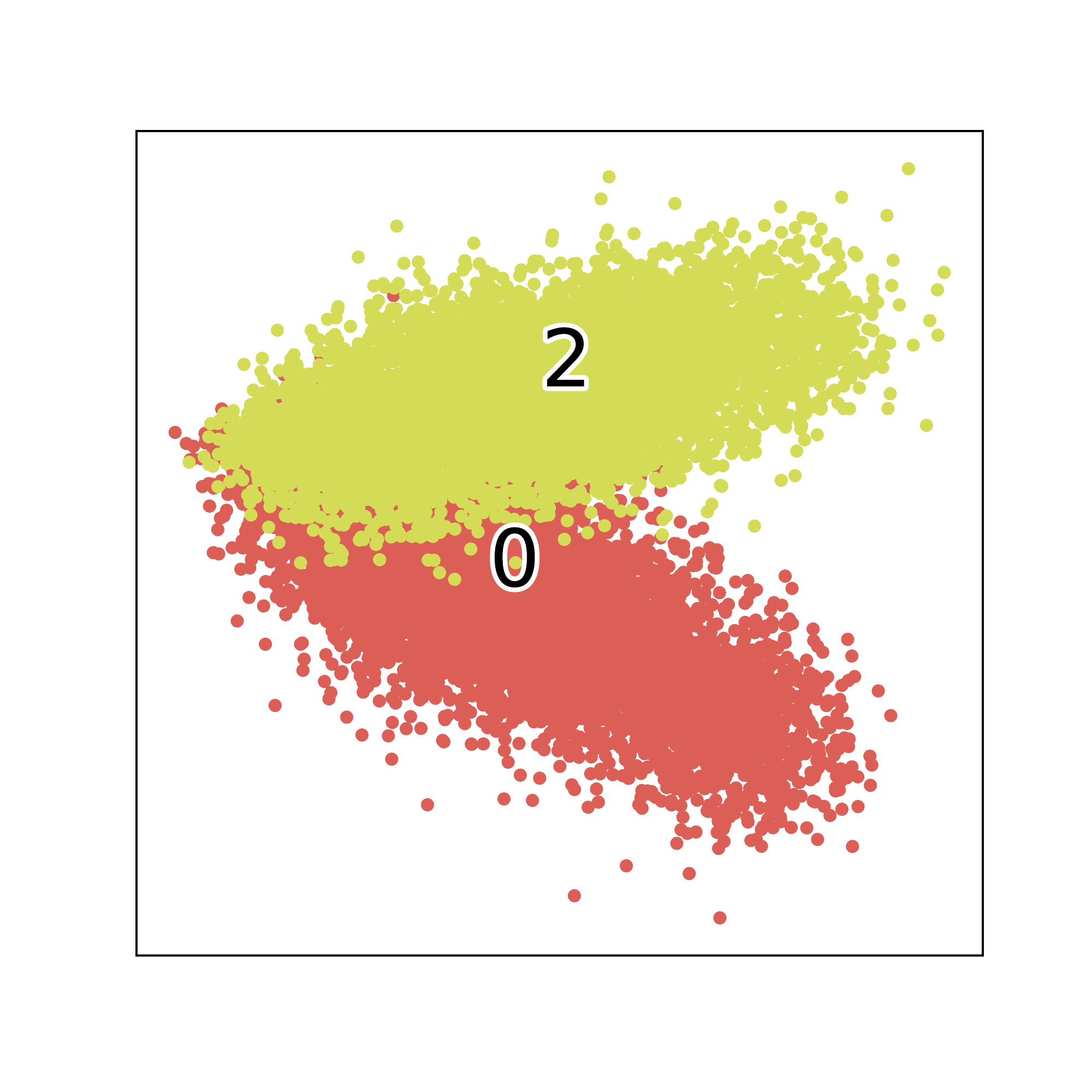}
	\caption{PCA of a bilingual speaker embedding for each
          utterance in 7 hours of English (red parts) and 7
          hours of Spanish (yellow parts).}
	\label{fig:enmx}
\end{figure}

\subsection{Bilingual speaker embedding}
The bilingual voice (Voice IDs 0 and 2 in in Table \ref{tab:8sp})
is used as a \emph{teacher} from which other voices can learn
to be bilingual in English and Spanish. 

The speaker encoder generates a speaker embedding for each utterance.
The speaker embedding space is trained to discriminate between
different speakers so that embeddings from the same speaker
for multiple utterances have high cosine similarity between each other 
compared to different speakers. Such discriminative training results in each
speaker utterances forming a cluster in the embedding space. We can observe 
such speaker clusters with PCA~\cite{wold1987principal} or tSNE~\cite{maaten2008visualizing} on the multiple speaker embedding 
space (One such example can be found in this work \cite{jia2018tarnsfer}).

However, the speaker encoder is generally trained
with speakers speaking only one language. There were no bilingual
speakers in our training data for the speaker encoder. 
This is mainly because, speaker encoders are trained with hundreds of speakers, and
it is difficult to get data from many bilingual speakers.  Thus, it
is interesting to learn how the speaker embedding space is
distributed when the same speaker is speaking in two different
languages, specifically whether it forms one cluster for both languages or two separate clusters. 
 To the best of our knowledge there has been no
previous work exploring such bilingual speaker embeddings. Since the speaker embeddings are used
to condition for speaker output, it is also important to know how language is represented in these embeddings.    

We explore the bilingual speaker embeddings from 14 hours of recorded speech (7 hours in Spanish and 7 hours in English) from the reference speaker.
First we visualize the embeddings with PCA and in Section \ref{sec:lda} we fit two Gaussian
distributions and check the accuracy of the fit.

In Figure \ref{fig:enmx} we
visualize language representation in bilingual speaker
embedding space with principal component analysis (PCA)
\cite{wold1987principal}. Our observations are:
\begin{enumerate}
\item These two languages have distinct clusters, and
\item these two language clusters have overlapping
  regions and then they diverge.  Further
  investigation shows that the overlapping is due to
  smaller sentences (fewer than five words).
\end{enumerate}

We hypothesize that any bilingual speaker space will have a
similar structure, i.e., two distinct but overlapping
clusters. If the monolingual voices could speak another
language we would expect them to have similar speaker
embedding clusters and so it should be feasible to modify them
to speak another language simply by translating the speaker
space, i.e. shifting the cluster corresponding to
one language towards that of another. We detail this procedure
in the following section.

\subsection{Speaker embedding translation}
\label{sec:speaker-embedding-translation}

We measure the mean
speaker embedding cluster as $\mu_{x}$, where $x$ is voice ID and 8
voices (6 + reference speaker in two languages) are available. The bilingual reference speaker has two
language cluster means $\mu_{r}^{A}$
and $\mu_{r}^{B}$ for English and Spanish respectively in the speaker
embedding space.  We compute the modification $\Delta_{A2B}$ needed to
convert between the average cluster embeddings in both
languages as:

\begin{equation}
\Delta_{A2B} = \mu_r^{A}  - \mu_r^{B}
\label{eq:one}
\end{equation}
Note that since the speaker is the same, this shift can only contain
information about the languages. Then for any different speaker $x$,
who only speaks language A, we can obtain their embedding in language B
as,

\begin{equation}
{\mu}_x^B = \mu_x^A + \epsilon \Delta_{A2B}
\label{eq:two}
\end{equation}
where $\epsilon$ $(0 \le \epsilon \le 1)$ is a scaling factor to
decide the level of the transformation, that is the level of native
accent to be expected with the converted embedding.  In our
experiments it is set to zero or one except for our experiments
investigating accent modification.

\section{Experiments}
\subsection{Dataset}
We use an internal 56 hour dataset of 8 studio recorded
voices, with $\sim 7$ hours of speech from each voice (counting the reference speaker as two voices with a total of $\sim 14$ hours of recordings). 
The  dataset is balanced in the two languages (four English / four
Spanish) and is balanced by gender, four male / four female. Two locales
are represented for each language: en-US and en-AU for English
and es-ES and es-MX for Spanish. For testing we
use total of 400 sentences in English and 400 in Spanish; the sentences
are $3-15$ words long. The reference speaker is
bilingual; all other speakers are monolingual. Using Eq. \ref{eq:two} we transfer any monolingual voices to another language.

\subsection{Linear Discriminant Analysis}\label{sec:lda}
As our first experiment, we fit two Gaussian distributions to the bilingual speaker embeddings from the recorded speech
with Linear Discriminant Analysis (LDA). We divide the
bilingual speech as $75\%$ / $25\%$ for train-test. This gives
12525 training and 4175 test sentences. Next, we fit an LDA
on training data to maximize class separability between English
and Spanish bilingual embeddings. On the test set we observe 99\%
accuracy. This further shows that the two languages of the
same speaker can be discriminated with very high accuracy.

\subsection{Listening Tests}

We carried out five listening tests to gauge the
subjective performance of our system. The tests fall into 3
categories. The first category is overall TTS naturalness (two
tests, one per language); the second examines cross-language voice similarity
(one test); the third examines the effect of accent on overall
TTS naturalness (two tests, for two language).

{\bf Naturalness:} A MOS test (Test 1) compared six variations of
synthesis of English, three trained from English recordings
(including a concatenative synthesis baseline
reference) and three using Spanish voices modified to
speak English (see Table \ref{tab:expt1}). 12 test
sentences were synthesized and played to 30 listeners,
who were native speakers of English. Listeners were
asked to rate voice naturalness on a 5-point scale
from (1) Bad to (5) Excellent. A second similar
experiment, Test 2, reversed the roles of Spanish and
English (see Table \ref{tab:expt2}).  

{\bf Similarity:} A similarity experiment (Test 3)
examined to what extent speech in a different language
can be identified as being from the same speaker. 84
pairs of sentences, with one sentence in English and
one in Spanish, using seven different voice combinations
(see Table \ref{tab:expt3}) were presented to 30 native
speakers of English and 30 native speakers of
Spanish. Listeners were asked to rate the voice
similarity for each pair of sentences on a scale from
(1) Very Different to (5) Very Similar.

{\bf Accent:} A MOS test (Test 4) asked listeners to
give a naturalness rating on a 5-point scale from (1)
Bad to (5) Excellent to English sentences generated
from 3 Spanish voices for different values of $\epsilon$
described in Equation \ref{eq:two}. A second MOS test (Test
5) reversed the roles of English and Spanish. Each
test had ten sentences and nine voice configurations, and
each test part had both English (30) and Spanish (30)
listeners.

\section{Results}

The naturalness results for Test 1 are shown in Table
\ref{tab:expt1}. The best scores are for Voice 1 and both versions
of the reference bilingual speaker. An ANOVA followed by a Tukey post-hoc test found
no statistically significant difference between these three
versions. Speaker 3 and speaker 6 fall into a lower-scoring second
group. The mapped version of the reference speaker performs
as well as the in-language version, an indication that the technique performs well.

\begin{table}[bt]
	\centering
	\footnotesize
	\begin{tabular}{| l | c | c | c |}
		\hline
		& Voice ID & MOS & Std. Dev.\\
		\hline
	        English          & $1$        & 4.06 & 0.88 \\
		Speakers         & $0$ ($ref_{en}$)        & 4.03 & 0.81 \\
		                 & baseline   & 2.48 & 1.16 \\
		\hline
		Spanish          & $3^{en}$   & 3.39 & 0.94 \\
		Speakers         & $2^{en}$   & 4.08 & 0.73 \\
		                 & $6^{en}$   & 3.32 & 0.88 \\
		\hline
	\end{tabular}
	\caption{MOS naturalness scores with English sentences. Lower section displays cross-lingual transfer following Eq. \ref{eq:two}.}
	\label{tab:expt1}
\end{table}

The results for Test 2 are shown in Table \ref{tab:expt2}.
For this experiment the scores for speaker 1 and both versions
of the reference bilingual speaker again showed no statistical
difference. Speakers 1, 3 and 4 formed a second group and finally
speaker ($0^{es}$) and speaker 3 were grouped together.  We
highlight that the cross-language version of speaker 1,
where the voice recordings are all for American English, was
rated significantly higher speaking Mexican Spanish than speaker 3,
a voice custom built to speak Mexican Spanish.  Taking the two
experiments together it seems likely that some element of
voice preference is factored into the results. We intend to
carry out further experiments to control for this effect.

\begin{table}[bt]
	\centering
	\footnotesize
	\begin{tabular}{| l | c |  c | c |}
		\hline
		& Voice ID & MOS & Std. Dev.\\
		\hline
		Spanish          & $3$        & 4.14 & 0.93 \\
		Speakers         & $2$ ($ref_{es}$)        & 4.34 & 0.81 \\
		                 & baseline   & 2.29 & 0.91 \\
		\hline
		English          & $0^{es}$   & 4.30 & 0.85 \\
		Speakers         & $1^{es}$   & 4.19 & 0.85 \\
		                 & $4^{es}$   & 4.10 & 0.92 \\
		\hline
	\end{tabular}
	\caption{MOS naturalness scores with Spanish sentences. Lower section displays cross-lingual transfer following Eq. \ref{eq:two}.}
	\label{tab:expt2}
\end{table}

The similarity results for Test 3 are shown in Table
\ref{tab:expt3}.  The higher the score the higher the perceived
similarity between pairs of voices in the two languages. Matched
voices with cross-lingual transfer scored higher. Pairs $3-3^{en}$
and $6-6^{en}$ scored highest, showing the model was
performing better for Spanish to English cross-lingual
transfer than for English to Spanish.
The two mixed voice combinations were rated
lower than the same-voice combinations and the difference was
statistically significant ($p<0.001$).  The bilingual speaker
was perceived as less similar than some combinations, probably
because the original recordings in each language had slightly
different style requirements.

\begin{table}[bt]
	\centering
	\footnotesize

	\begin{tabular}{| l | c | c |}
		\hline
		Voice ID pairs               & MOS & Std. Dev.  \\
		\hline                       
		$3^{en}-3$                   &  3.99 & 1.08 \\
		$6^{en}-6$                   &  3.84 & 1.22 \\
		$1-1^{es}$                   &  3.37 & 1.28 \\
		$4-4^{es}$                   &  3.16 & 1.26 \\
		$0-2$ ($ref_{en}-ref_{es}$)    &  3.05 & 1.30 \\               %
		$1-3^{es}$                   & 2.50 & 1.22 \\
		$0-6^{es}$ ($ref_{en}-6^{es}$) & 2.43 & 1.27 \\
		\hline
	\end{tabular}
	\caption{MOS similarity scores - English and Spanish. }
	\label{tab:expt3}
\end{table}

The cross-lingual transfer accent control results for Tests 4 and 5 are shown in Tables
\ref{tab:expt4} and \ref{tab:expt5} respectively. Generally
accent seems to be controllable and the least-accent cases are
close to accent-free. For experiment 4, there was no
statistically significant difference in terms of naturalness
between the 3 reference speaker variants, nor the 3 speaker 3 variants, but
the $\epsilon=0$ version of speaker 6 was rated significantly
lower.

For experiment 5, the results again show no significant
difference for the reference speaker variants, but for the other
two voices all the rating differences were significant.  We
need to investigate further to try to differentiate accent
preference from naturalness.

\begin{table}[bt]
	\centering
	\footnotesize
	
	\begin{tabular}{| l | c c c |}
		\hline
		Voice  ID & $\epsilon= 0$ & $\epsilon= 0.5$ & $\epsilon= 1$ \\
		\hline
		$2_\epsilon^{en}$  & 4.08  & 4.22  & 4.16 \\
		$3_\epsilon^{en}$  & 3.64  & 3.76  & 3.76 \\
		$6_\epsilon^{en}$  & 3.43  & 3.69  & 3.73 \\
		\hline
	\end{tabular}
	\caption{MOS quality with foreign accent - English}
	\label{tab:expt4}
\end{table}

\begin{table}[bt]
	\centering
	\footnotesize

	\begin{tabular}{| l | c c c |}
		\hline
		Voice  ID & $\epsilon= 0$ & $\epsilon= 0.5$ & $\epsilon= 1$ \\
		\hline
		$0_\epsilon^{es}$ &  3.95  &  4.23  &  4.22  \\
		$1_\epsilon^{es}$ &  3.42  &  3.89  &  4.16  \\
		$4_\epsilon^{es}$ &  3.35  &  3.86  &  4.16  \\
		\hline
	\end{tabular}
	\caption{MOS quality with foreign accent - Spanish}
	\label{tab:expt5}
\end{table}

	\begin{figure}[htb!]
	\centering
	\includegraphics[width=0.6\linewidth,  trim={3cm 3cm 2.5cm 3cm},clip]{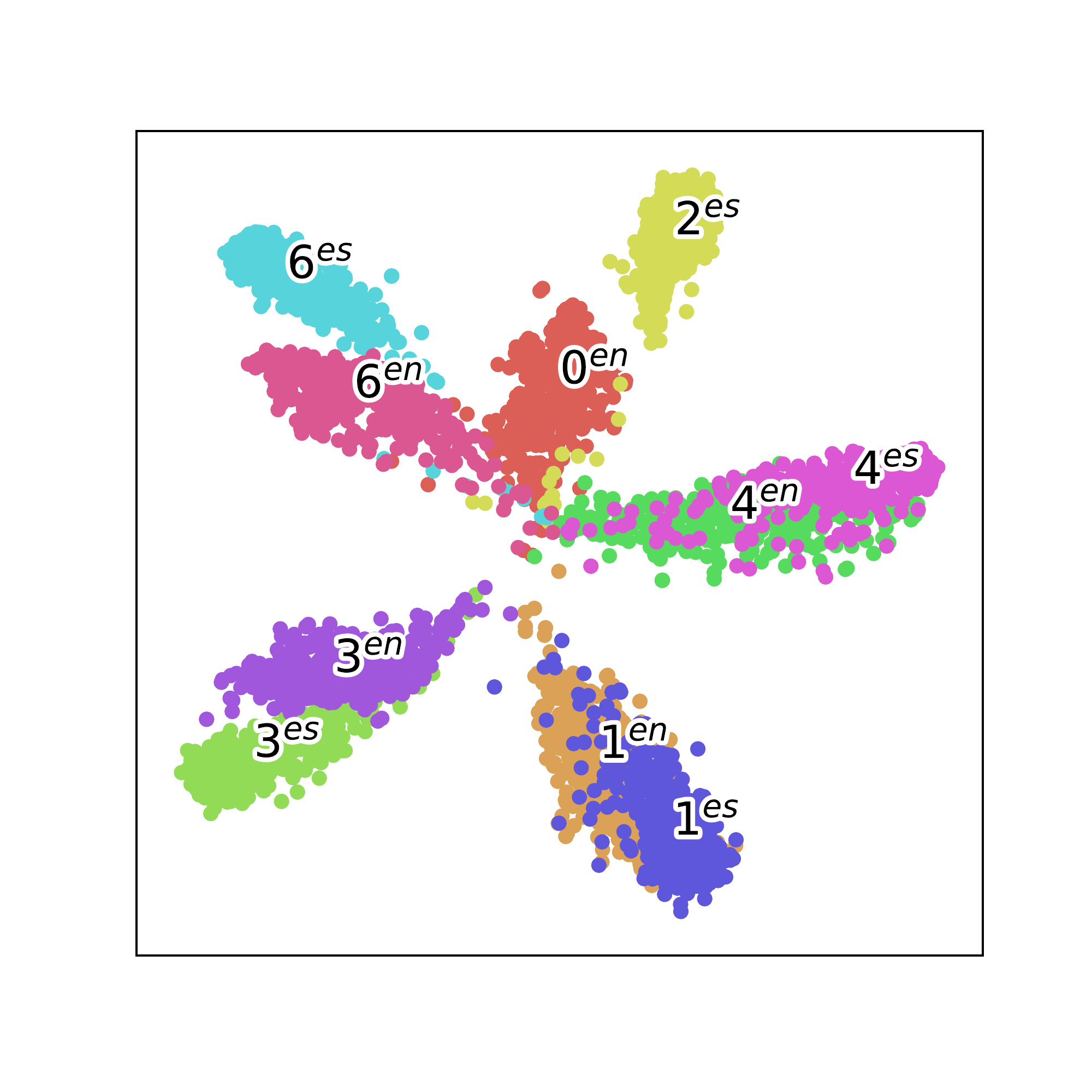}
	\caption{ tSNE plot with speaker embedding with
          synthesized speech. Monolingual speakers:
          $0^{en}$, $1^{en}$, $4^{en}$,
          $2^{es}$, $3^{es}$, $6^{es}$. Cross-lingual
          speakers: ${3}^{en}$, ${6}^{en}$,
          ${1}^{es}$, ${4}^{es}$. Same speaker English-Spanish
          clusters after cross-lingual transfer are close-by.  }
	\label{fig:synth}
\end{figure}

\section{Visualizing effects of cross-lingual transfer in synthesized speech}
Finally, on synthesized speech we visualize the effect of
speaker space translation: i.e. whether the TTS synthesized
speech also mimics the bilingual speaker distribution or
not. For this test we have 400 English and 400 Spanish
sentences.

Synthesized test sentences are generated as follows: For the
bilingual speaker we synthesize with their embeddings, English
sentences with mean English speaker embedding and Spanish with mean
Spanish speaker embedding. For monolingual speakers, we
synthesize their native language with no translation and
cross-lingual transfer with translation.  So this gives us five
speakers speaking two languages, one bilingual speaker and two English
and two Spanish speakers speaking both languages, 10 voices in
total.  We visualize with a tSNE~\cite{maaten2008visualizing} plot in Figure
\ref{fig:synth}. Here we can see that 10 voices form five
distinct clusters and each cluster contains two
subclusters. Here five clusters represent five speaker and two
subclusters represent two languages. From the
visualization it is easy to see speaker clusters after
cross-lingual transfer are close-by and overlapping with their
native speech clusters, following the structure similar to the
bilingual speaker. Hence, speaker identity is maintained
through cross-lingual transfer.

\section{Conclusions}

The formalism we have developed leads to high quality TTS in a
second language without losing the characteristics of the
voice. We found that there is a clustering by language in
speaker embedding space for a bilingual speaker and we were
able to use the cluster means to help control language and
accent at inference time.  There are a number of reasons why
this technique is extremely interesting: (1) One key point is
that in the transformation process the quality remains high.
(2) It requires a relatively modest amount of data. (3) Having
data from one bilingual speaker helps make other monolingual
speaker bilingual, without the complexities of trying to
record a monolingual speaker speak a second language. (4) We
cite particularly as evidence of the promise of the technique
that in one case we demonstrated a transformed voice that
performed better than a high quality in-language voice.  (5)
It is possible to control the degree of accent present in the
synthesis. All these are very desirable characteristics for
synthesis.

\newpage
\bibliographystyle{IEEEbib}
\bibliography{strings,refs}
	
\end{document}